\PassOptionsToPackage{dvipsnames}{xcolor}
\documentclass{ieeetmimod}
\ifdefined\showrevision

\usepackage{xcolor}
\usepackage[normalem]{ulem}
\usepackage{marginnote}
\usepackage{xspace}

\newcommand{\revref}[2]{%
\marginnote{$R_{#1}C_{#2}$}
}

\newcommand{\edtref}[1]{%
\marginnote{$EC_{#1}$}
}

\newcommand{\revdel}[1]{%
{\color{darkgray}\sout{#1}}\xspace
}

\else

\usepackage{xspace}

\newcommand{\marginnote}[1]{\ignorespaces}
\newcommand{\revref}[2]{\ignorespaces}
\newcommand{\edtref}[1]{\ignorespaces}

\newcommand{\revdel}[1]{\ignorespaces}

\fi
\usepackage{amsfonts}
\usepackage{bm}
\usepackage{placeins}
\usepackage{multirow}
\usepackage{algorithm}
\usepackage{algpseudocode}
\usepackage{threeparttable}
\usepackage{amssymb}
\usepackage{amsmath} 
\usepackage{wasysym}
\DeclareFontFamily{U}{wasy}{}
\DeclareFontShape{U}{wasy}{m}{n}{
     <-5.5> wasy5
  <5.5-6.5> wasy6
  <6.5-7.5> wasy7
  <7.5-8.5> wasy8
  <8.5-9.5> wasy9
     <9.5-> wasy10
}{}
\DeclareFontShape{U}{wasy}{b}{n}{
 <-10> ssub * wasy/m/n
 <10-> wasyb10
 }{}
\DeclareFontShape{U}{wasy}{bx}{n}{ <-> ssub * wasy/b/n}{}
\DeclareFontShape{U}{wasy}{m}{sl}{ <-> wasysl10 }{}
\DeclareFontShape{U}{wasy}{m}{it}{ <-> ssub * wasy/m/sl }{}

\usepackage{tikz}

\newcommand{\diamondrightblack}{%
  \tikz[scale=0.11, baseline=-0.6ex]{
    \draw[line width=0.4pt] (0,0) -- (1,1) -- (2,0) -- (1,-1) -- cycle;

    \path[fill=black] (2,0) -- (0.95,1) -- (0.95,-1) -- cycle;
  }%
}

\usepackage[dvipsnames]{xcolor}
\usepackage[english]{babel}
\usepackage{booktabs}
\usepackage{subcaption}
\usepackage{xspace}

\newcommand{\sepsymb}{\rule{0.14em}{\dimexpr 0.8\ht\strutbox}\hspace{0.2em}\rule{0.14em}{\dimexpr 0.8\ht\strutbox}}
\newcommand{\tgtsymb}{\rule{0.6em}{\dimexpr 0.8\ht\strutbox}}

\newcommand{\ie}{\textit{i.e.}\xspace}

\usepackage{glossaries-extra}
\setabbreviationstyle[acronym]{long-short}
\newacronym{iis}{IIS}{instrument-level instance segmentation}
\newacronym{iss}{ISS}{instrument-level semantic segmentation}
\newacronym{pss}{PSS}{part-level semantic segmentation}
\newacronym{pis}{PIS}{part-aware instance segmentation}

\usepackage{cite}
\usepackage{graphicx}
\usepackage{textcomp}
\definecolor{githubpink}{HTML}{D63384}
\colorlet{subsectioncolor}{black}
\hypersetup{colorlinks=true,allcolors=black,urlcolor=githubpink}
\patchcmd{\keywords}{Index Terms}{Keywords}{}{\errmessage{Keyword heading patch failed}}
\patchcmd{\keywords}{Index Terms}{Keywords}{}{\errmessage{Keyword heading patch failed}}
\patchcmd{\keywords}{Index Terms}{Keywords}{}{\errmessage{Keyword heading patch failed}}

\begin{document}
\bstctlcite{IEEEexample:BSTcontrol}

\title{SurgPIS: Surgical-instrument-level Instances and Part-level Semantics for Weakly-supervised Part-aware Instance Segmentation}
\author{
Meng Wei, Charlie Budd, Oluwatosin Alabi, Miaojing Shi, \IEEEmembership{Senior Member, IEEE}, and Tom Vercauteren
\thanks{Meng Wei is supported by the UKRI EPSRC CDT in Smart Medical Imaging [EP/S022104/1]. This work was supported by core funding from the Wellcome Trust / EPSRC [WT203148/Z/16/Z; NS/A000049/1].  Miaojing Shi is supported by the Xiaomi Young Scholar Project and China Fundamental Research Funds for the Central Universities. 
For the purpose of open access, the authors have applied a CC BY public copyright licence to any Author Accepted Manuscript version arising from this submission. Tom Vercauteren is a co-founder and shareholder of Hypervision Surgical.}
\thanks{Meng Wei, Charlie Budd, Oluwatosin Alabi, and Tom Vercauteren are with the School of Biomedical Engineering and Imaging
Sciences, King’s College London, WC2R 2LS London, U.K. (e-mail: meng.wei@kcl.ac.uk; charles.budd@kcl.ac.uk; oluwatosin.alabi@kcl.ac.uk; tom.vercauteren@kcl.ac.uk).}
\thanks{Miaojing Shi is with the College of Electronic and Information Engineering, Tongji University, 201804 Shanghai, China. (e-mail: mshi@tongji.edu.cn).}
\thanks{Corresponding author: Miaojing Shi}
}

\maketitle
\thispagestyle{plain}

\begin{abstract}
Consistent surgical instrument segmentation is critical for automation in robot-assisted surgery.
Yet, existing methods only treat \gls{iis} or \gls{pss} separately, without interaction between these tasks. 
In this work, we formulate surgical tool segmentation as a unified \gls{pis} problem and introduce SurgPIS, 
the first end-to-end \gls{pis} model for surgical instruments.
Our method adopts a transformer-based mask classification approach 
and introduces part-specific queries derived from instrument-level object queries, explicitly linking parts to their parent instrument instances. 
In order to address the lack of large-scale datasets with both instance- and part-level labels, we propose a weakly-supervised learning strategy for SurgPIS to learn from disjoint datasets labelled for either \gls{iis} or \gls{pss} purposes. 
During training, we aggregate our \gls{pis} predictions into \gls{iis} or \gls{pss} masks, thereby allowing us to compute a loss against partially labelled datasets.
A student-teacher approach is developed to maintain prediction consistency for missing \gls{pis} information in the partially labelled data, e.g. parts for \gls{iis} labelled data.
Extensive experiments across multiple datasets validate the effectiveness of SurgPIS, achieving state-of-the-art performance in \gls{pis} and superior results in \gls{iis}, \gls{pss}, and instrument-level semantic segmentation.
The code and weights are available at \url{https://github.com/weimengmeng1999/SurgPIS}.

\end{abstract}

\begin{IEEEkeywords}
Robotic-assisted surgery, Surgical vision, Surgical instrument segmentation, Part-aware instance segmentation, Weakly-supervised learning
\end{IEEEkeywords}

\section{Introduction}
Surgical instrument segmentation is crucial for automation in robot-assisted minimally invasive surgery~\cite{palep2009robotic,allan20202018}.
Different levels of granularity are needed for different downstream tasks.
Binary tool segmentation may for example allow for improved augmented-reality overlays \cite{allan20202018}, 
per-instance identification may enable pose estimation~\cite{sestini2021kinematic}, 
and instrument part identification may help in understanding instrument-tissue interactions~\cite{nwoye2020recognition}. 
A unified model that can detect and segment instruments at different granularity levels is a foundation to support downstream tasks such as tool tracking \cite{cheng2021deep}, trajectory prediction~\cite{toussaint2021co}, surgical action triplets recognition~\cite{nwoye2023cholectriplet2021}, and higher-level autonomous robotic surgery
\cite{lu2021toward}, paving the way for next-generation of AI assistance in the operating room.

\begin{figure}[t!]
  \centering
  \includegraphics[width=\linewidth]{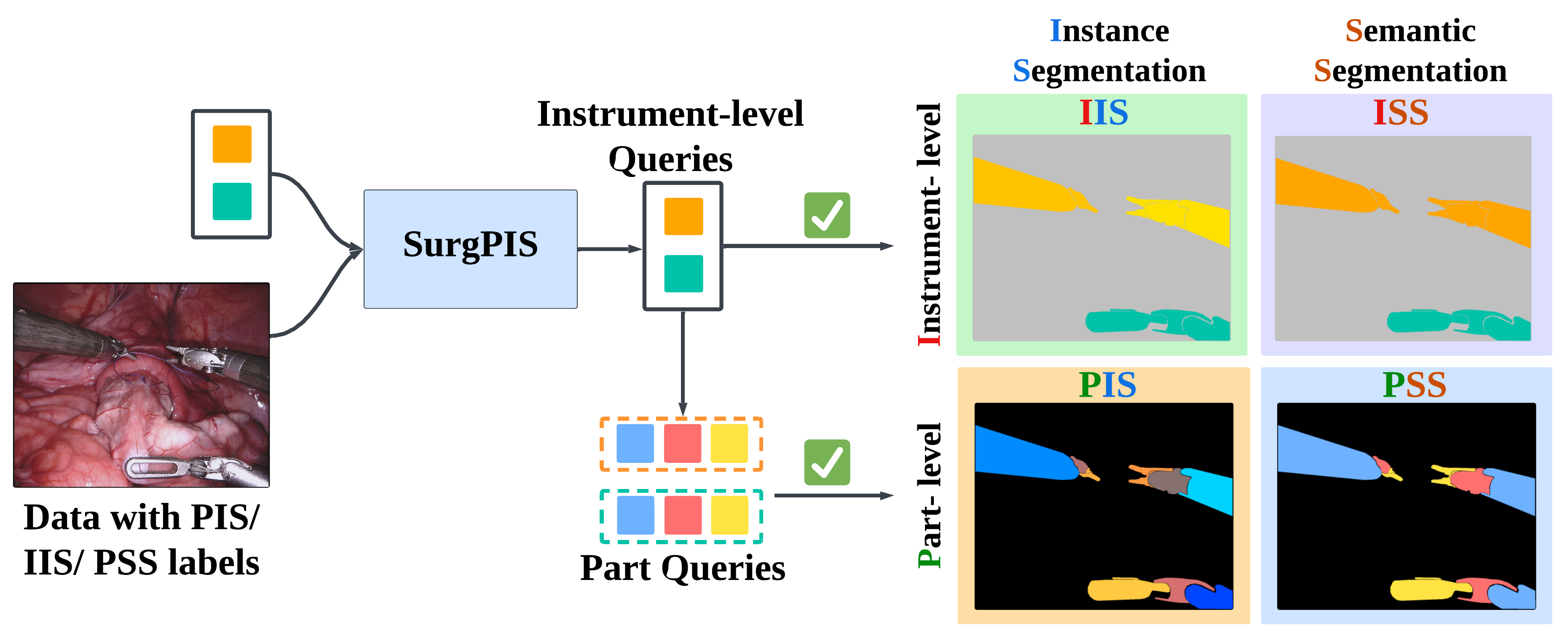}
  \caption{\textit{SurgPIS} is the first unified model for surgical instrument segmentation, capable of predicting both instrument-level instances (\ie IIS and ISS) and part-aware instances (\ie PIS and PSS) based on a query-based transformation approach. It leverages weak PIS supervision from disjoint PSS and ISS datasets to learn from partially labelled data in different granularities.\label{fig:graphabstr}}
\end{figure}
Most existing methods treat instrument segmentation as a mere multi-class semantic segmentation problem, focusing either on instrument types~\cite{jin2019incorporating,zhoutext2023,wei2024enhancing}, 
referred to as \glsfirst{iss}, or on part types~\cite{allan20192017,allan20202018,shvets2018automatic,psychogyios2023sar}, known as part-level semantic segmentation (\gls{pss}),
and very rarely both~\cite{shvets2018automatic}. 
Some studies~\cite{gonzalez2020isinet,zhao2022trasetr,baby2023forks} have explored \glsfirst{iis} to  
capture individual instruments. No existing approach unifies the aforementioned tasks.
In this work, we address surgical instrument segmentation as a \glsfirst{pis} task, which simultaneously distinguishes instrument instances and their composing parts. Our proposed SurgPIS approach, illustrated in \figref{fig:graphabstr}, is to the best of our knowledge the first \gls{pis} model proposed for surgical scene understanding. While existing methods can technically perform \gls{pis} by combining the output of different decoders or separate models, SurgPIS addresses a fundamental gap by integrating multi-granularity segmentation into a single model, enabling superior performance, efficient learning from disjoint datasets, and structured representations critical for advanced surgical applications.

Current part-aware panoptic segmentation methods for natural images~\cite{fang2021instances,du2018articulated,de2024task}, typically built upon Mask2Former~\cite{cheng2022masked}, either use separate learnable queries for object-level and part-level segmentation~\cite{fang2021instances,du2018articulated} or employ shared queries to jointly predict objects and parts~\cite{de2024task}.
SurgPIS adopts the shared query approach \cite{de2024task} but introduce a key novelty, part-specific query transformation, to explicitly build hierarchical links between parts and their parent surgical instrument instances, ensuring a structured representation that enables multi-granularity reasoning. 

Surgical instrument segmentation presents unique challenges beyond natural images due to the lack of comprehensive large-scale datasets. Existing large-scale datasets provide either only annotations for \gls{pss}~\cite{allan20192017,allan20202018,psychogyios2023sar} or only for \gls{iis}~\cite{ross2020robust,alabi2024cholecinstanceseg}, thereby limiting their use for \gls{pis} tasks. 
To address this, we introduce a novel weakly-supervised training pipeline for SurgPIS, leveraging a teacher-student framework to bridge the gap between datasets providing only \gls{pss} or \gls{iis} labels.
Our key innovation lies in aggregating the \gls{pis} predictions into \gls{pss} or \gls{iis} masks to enable
a straightforward loss computation for partially labelled datasets.
Concurrently, the teacher ensures \gls{pis} consistency between the teacher’s prediction and the student’s prediction under different augmentations for the same image.
This structured learning approach enhances \gls{pis} by leveraging information across different annotation granularities, enabling knowledge transfer from disjoint datasets with complementary labels (either \gls{pss}-only or \gls{iis}-only), a capability absent in existing part-aware panoptic segmentation models~\cite{li2022panoptic, li2024panoptic, de2024task} and surgical instrument segmentation methods~\cite{gonzalez2020isinet,
yue2024surgicalsam,ayobi2024pixel}.

\textbf{Contributions:} SurgPIS is the first \gls{pis} model for surgical instruments. We propose a novel weakly-supervised learning strategy to enable training on disjoint datasets lacking \gls{pis} labels. 
Extensive experiments first demonstrate the effectiveness of SurgPIS with EndoVis2018~\cite{allan20202018}, which provides complete \gls{pis} annotations.
We then validate our weakly supervised training strategy with EndoVis2017~\cite{allan20192017}, which contains only \gls{iis} labels and SAR-RARP50~\cite{psychogyios2023sar}, which includes only \gls{pss} labels.
We also evaluate our model on GraSP~\cite{ayobi2024pixel}, which we annotated for \gls{pis} and acts as a dataset unseen during training.
Evaluations on \gls{pss}, \gls{iis}, and \gls{iss} tasks, alongside comparisons with existing methods, highlight the adaptability of SurgPIS across diverse surgical instrument segmentation tasks.

\section{Related Work}
\subsection{Surgical Instrument Segmentation}
Existing surgical instrument segmentation approaches predominantly focus on either \gls{iis} tasks or semantic segmentation tasks, including \gls{pss} and \gls{iss}.
For \gls{iis} tasks \cite{alabi2024cholecinstanceseg}, ISINet~\cite{gonzalez2020isinet} introduces an instance-based segmentation approach that leverages a temporal consistency module to detect instrument candidates. TraSeTR \cite{zhao2022trasetr} leverages tracking cues within a Track-to-Segment transformer framework to improve instance-level instrument segmentation.  
For semantic segmentation tasks \cite{jin2019incorporating,ayobi2023matis,zhoutext2023,wei2024enhancing}, methods such as MF-TapNet \cite{jin2019incorporating} utilize motion flow information into an attention pyramid network, training separate models for part and instrument semantic segmentation. 
MATIS \cite{ayobi2023matis} employs a two-stage transformer framework, first generating fine-grained instrument region proposals with masked attention modules, then refining and classifying them using pixel-wise attention for instance-level predictions in the second stage.
While these methods achieve strong performance in their respective tasks, they can not be directly extended for the \gls{pis} task.
Multi-task methods such as TAPIS \cite{ayobi2024pixel} can only achieve \gls{pis} through post-processing by combining outputs from task-specific decoders and dealing with any inconsistencies between the subtask outputs.
This post-hoc fusion inherently fails to establish an explicit hierarchy between an instrument instance and its constituent parts.
Our work addresses this gap by introducing a unified approach for part-aware instance segmentation.  

\subsection{Part-aware Panoptic Segmentation}
Part-aware panoptic segmentation \cite{de2021part} extends panoptic segmentation \cite{kirillov2019panopticseg} by incorporating part-level segmentation within object-level segments. 
When applied to datasets containing only instance-level objects ("things") but no semantic-only ("stuff") classes, part-aware panoptic segmentation reduces to \gls{pis}.
Each instance with parts is explicitly decomposed into its constituent parts.
This strategy has been widely used in human parsing \cite{zhang2022aiparsing,wang2023contextual} and found some initial applications in biological cell parsing~\cite{chen2024cp}. 
Existing \gls{pis} methods typically use separate models \cite{zhang2022aiparsing,wang2023contextual} or distinct segmentation heads \cite{chen2024cp} for object instance and part segmentation, later fusing the results.  
Similar to the PIS case, most part-aware panoptic segmentation approaches do not directly predict parts per object instance but instead generate separate panoptic and part-segmentation predictions \cite{jagadeesh2022multi,li2022panoptic,li2024panoptic}. 
%
JPPF \cite{jagadeesh2022multi} uses a shared encoder with separate heads for semantic, instance, and part segmentation, combining their outputs through a rule-based fusion module to produce consistent panoptic-part predictions.
Panoptic-PartFormer \cite{li2022panoptic} introduces a Transformer-based framework with separate learnable queries for thing, stuff, and part segments, employing a decoupled decoder for initial mask predictions and a cascaded Transformer decoder for query-based reasoning. 
Panoptic-PartFormer++ \cite{li2024panoptic} refines this by incorporating cross-attention modules to enhance local part feature interactions.  

Moving away from treating instances and parts separately,
TAPPS \cite{de2024task} proposes a novel approach to leverage joint object-part representations.
The network utilizes a set of shared queries to jointly predict object-level segments and their corresponding part-level segments, significantly outperforming methods that separately predict objects and parts. 
However, as shown in our subsequent experiments, TAPPS fails to transfer directly to the surgical domain.
Different instrument categories indeed often have similar appearances \cite{ayobi2023matis} and identical part types (e.g. all surgical instruments have a shaft), unlike natural images where object classes typically consist of distinct part types (e.g. cars do not have limbs).We address the surgical domain challenge by explicitly building the hierarchy between instrument instances and their identical part types and incorporating part-level loss to enhance model learning.

\subsection{Weakly-supervised Instance Segmentation}
Weakly supervised instance segmentation comprises a broad range of techniques depending on the available supervision signal.
Most previous works focus on consistent weak supervisory signals rather than disjoint datasets with partial annotations.
One approach leverages relaxed bounding box supervision~\cite{zhou2018weakly,tian2021boxinst,cheng2023boxteacher,cheng2022pointly}.
Another approach learns instance segmentation without direct mask annotations. 
DiscoBox~\cite{lan2021discobox} for example 
introduces a self-ensembling framework where a structured teacher refines masks, establishes dense correspondences, and generates pseudo-labels to supervise instance segmentation and semantic correspondence.
Point-supervised methods \cite{cheng2022pointly} have shown that instance segmentation models 
can be effectively adapted to point-based supervision, offering faster training and a more cost-efficient annotation process. 
There are also works based on semantic knowledge transfer~\cite{kim2022beyond} and continual learning for weakly-supervised semantic segmentation~\cite{hsieh2023class}. 
Given the unique data availability in the surgical domain, where most of the datasets are partially labelled by different segmentation tasks, a unique weakly-supervised training strategy that can leverage disjoint datasets is needed.
Previous work has proposed approaches to train semantic segmentation networks with disjoint datasets~\cite{dorent2021learning}.
However, the adaptation to instance segmentation is not straightforward. 
Our work enables learning \gls{pis} from disjoint surgical datasets with complementary annotations via an aggregation based weakly-supervised strategy.
\section{Method}
\label{sec:method}
We propose a two-stage training procedure for SurgPIS. In the first stage, our SurgPIS model is trained using a (potentially small) \gls{pis} labelled dataset $\mathcal{D}_{\textrm{PIS}}$, making this stage a fully supervised \gls{pis} task. In the second stage, we train our SurgPIS keeping the  \gls{pis} labelled data $\mathcal{D}_{\textrm{PIS}}$ but introducing weak supervision from disjoint datasets $\mathcal{D}_{\textrm{IIS}}$ and $\mathcal{D}_{\textrm{PSS}}$ labelled with only \gls{iis} or \gls{pss} labels respectively.

\subsection{Label Representation and Notations}
\label{sec:Notations}
Let $C^{\textrm{instr}}$ be the number of surgical instrument classes (e.g. Bipolar Forceps, Large Needle Driver etc.), excluding the background, and $C^{\textrm{inp}}$ be the input classes including tissue background and instrument classes i.e., $C^{\textrm{inp}} = C^{\textrm{instr}} + 1$.
Each instrument type can be decomposed into a maximum number $C^{\text{part}}$ of parts (e.g., shaft, wrist, clasper). 
We categorize our labelled data into three types. The first type consists of samples with \gls{pis} labels:
\begin{equation}
    \mathcal{D}_{\textrm{PIS}} = \Big\{(x, \big\{(y_0,c_0);(y_i,c_i,\{ m_{i,k}\}_{k<C^{\text{part}}})\big\}_{0<i<N_x})\Big\}
\label{eq:representation}
\end{equation}
where $x$ is an image of size $H \times W$ associated with $N_x$ instances;
$y_0$ is the binary instance mask for tissue background with associated one-hot encoded background class label $c_0$;
$y_i$ is a binary instance mask with associated one-hot encoded instrument class label $c_i$;
and $m_{i,k}$ is the part-level instance binary mask 
in a predefined and fixed order.
Consistency across the instrument and part masks implies that $y_i = \bigcup_k m_{i,k}$ and $m_{i,k} \cap m_{i,k'} = \emptyset$.
We also assume non-intersecting instrument masks: $y_{i} \cap y_{j} = \emptyset$.
%
The second type of labelled data $\mathcal{D}_{\textrm{IIS}}$ has the same instrument-level labels as \(\mathcal{D}_{\textrm{PIS}}\) 
and simply lacks the part masks.
The third type of labelled data $\mathcal{D}_{\textrm{PSS}}$ contains only 
semantic segmentation maps $\{s\}$ with \gls{pss} labels represented by one-hot encoding on a per-pixel basis:
$\mathcal{D}_{\textrm{PSS}} = \{(x,s)\}$.

The predictions from SurgPIS follow the same label representations as the training data in \eqref{eq:representation}, except for the fact that crisp binary masks are replaced by soft masks and one-hot encoded labels are replaced by probability vectors.
We use a hat notation and typically use $j$ for indexing to distinguish between ground truth and predictions, e.g. 
$\hat{y}_j \in [0;1]^{H\times W}$,
$\hat{c}_j \in \Delta^{C^{\textrm{inp}}+1}$ (probability vector including the tissue background class and “no
object” class\cite{cheng2022masked}),
and $\hat{s}_k \in \big(\Delta^{C^{\textrm{part}}+1}\big)^{H \times W}$.
We assume $\hat{c}_j[0]$ is the probability for tissue background class;
We also assume that the number of instrument instances $N_x$ is upper bounded by a fixed value $N_q$.

\subsection{SurgPIS Architecture}
\begin{figure*}[htb]
  \centering
  \includegraphics[width=\linewidth]{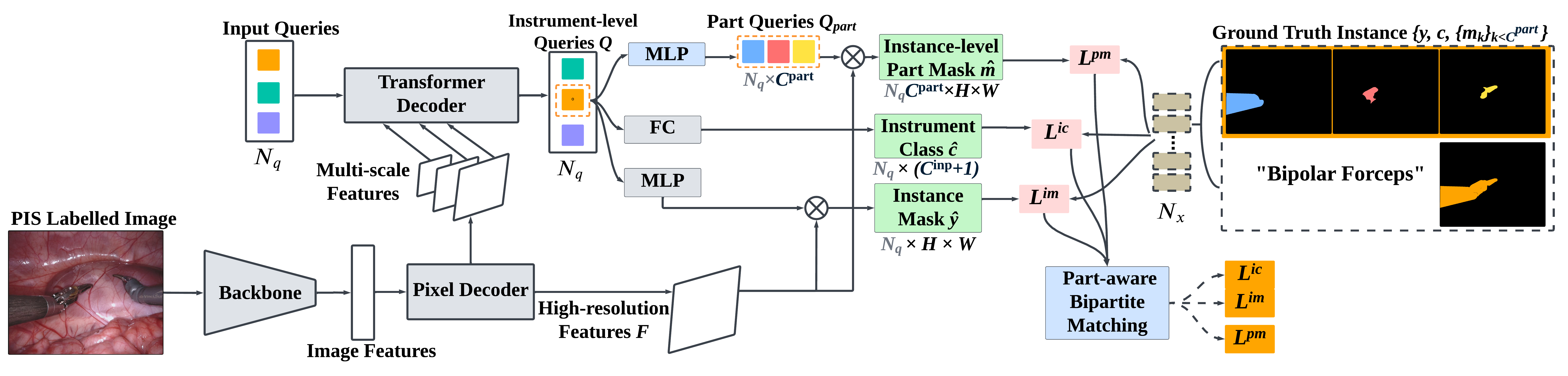}
  \caption{\textit{SurgPIS} extends Mask2Former (grey) by a novel query transformation module to get part-specific queries, and part-aware bipartite matching (blue) to calculate the losses at the instrument-level and part-level.\label{fig:fs}}
\end{figure*}
As shown in \figref{fig:fs}, SurgPIS extends Mask2Former~\cite{cheng2022masked} by transforming instrument queries to part-specific queries to predict instrument classes, instrument-level instance masks, and \gls{pis} masks. The instrument-level losses and part-level mask loss are supervised through hierarchical bipartite matching against \gls{pis} labels.

\subsubsection{Instrument-level Data Flow}
\label{sec:iisgen}
SurgPIS produces a high-resolution feature map $F \in \mathbb{R}^{C_\epsilon \times H \times W}$,
where $C_\epsilon$ is the embedding size,
by extracting multi-scale features using a backbone network (ResNet~\cite{he2016deep} or Swin Transformer~\cite{liu2021swin}) and refining them through a pixel decoder (Semantic FPN~\cite{kirillov2019panoptic,lin2017feature}).
The \( N_q\) learnable instrument-level queries are represented as \( Q \in \mathbb{R}^{N_q \times C_\epsilon}\).
Queries are derived by transforming a set of learnable positional embeddings using a standard transformer decoder which applies self-attention mechanisms across queries and cross-attention with the extracted image features.
The $j^{\textrm{th}}$ query encodes information about an instrument instance prediction or the background prediction and is used to predict its class label $\hat{c}_j$ and corresponding segmentation mask $\hat{y}_j$. 

\subsubsection{Part-specific Query Transformation}
\label{sec:pisgen}  
After obtaining the processed instrument-level queries, our key novelty lies in transforming instrument-level queries \( Q\), which encode global instrument-level information, into part-specific queries 
\( Q_{\text{part}} \). 
Unlike TAPPS \cite{de2024task}, which uses generic instrument-part-specific queries (separate queries for identical parts across instruments, e.g., 9 queries for 3 instruments with 3 parts), our part-specific queries are shared across instrument types (3 queries in total), which enables learning consistent part semantics.
To generate the part-specific queries $Q_{\text{part}} \in \mathbb{R}^{(C^{\text{part}} \times N_q ) \times C_\epsilon} $, we feed the instrument queries $Q$ 
to a multilayer perceptron (MLP).
To generate the part-specific binary mask predictions $\hat{m}_{j,k}$, we compute the product of $Q_{\text{part}}$ and the pixel-wise features in the high-resolution feature map $F$ and apply a sigmoid activation.

\subsubsection{Part-aware Bipartite Matching}
\label{sec:hbm}
To associate predictions with ground-truth instances, Mask2Former~\cite{cheng2022masked} and DETR~\cite{carion2020end} use a bipartite matching with a cost matrix that exploits a loss function between candidate pairs of masks. 
When \gls{pis} labels are available, we introduce a seemingly small but important change by constructing the cost matrix using not only an instrument-level class loss \(L^{\textrm{ic}}\) and instrument-level mask loss \(L^{\textrm{im}}\) but also a part-level mask loss \(L^{\textrm{pm}}\).
%
A generic mask loss \( \ell_M \) including focal loss and Dice loss \cite{cheng2022masked} is used 
for both types of masks.
The cross-entropy loss
\( \ell_{\text{CE}} \) 
is used for class predictions.
The instrument-level losses are defined as below with $i \in N_x$ and $j \in N_q$.
\begin{align}
   L^{\textrm{ic}}_{i,j} = \ell_{\text{CE}}(\hat{c}_j, c_i), ~~
    L^{\textrm{im}}_{i,j} = \ell_M(\hat{y}_j, y_i)
\end{align}
Given the fixed set and ordering of part classes, the part-level loss only contains a mask loss component averaged across the parts for instrument classes that exclude tissue background: 
\begin{align}
   L^{\textrm{pm}}_{i\neq 0,j} = \frac{1}{C^{\text{part}}} \sum_{k=1}^{C^{\text{part}}}\ell_M(\hat{m}_{j,k}, m_{i,k})
\end{align}
Following bipartite matching, a ground truth instance $i$  will be matched with a prediction instance that we denote a $\tau_i$.
To minimize memory requirements, we follow ~\cite{kirillov2020pointrend} and sample a limited number of points from the high-resolution features using importance sampling.

\subsubsection{Supervised Loss} 
\label{sec:suloss}
Given a set of matched instance indices $(i,\tau_i)$, the overall instrument-level class loss and masks loss can be calculated 
by averaging the contribution of each pair:
\begin{align}
   L^{\textrm{ic}}(\hat{c},c) &= \frac{1}{N_x} \sum_{i<N_x} \ell_{\text{CE}}(\hat{c}_{\tau_i},c_i) \\
   L^{\textrm{im}}(\hat{y},y) &= \frac{1}{N_x} \sum_{i<N_x} \ell_M(\hat{y}_{\tau_i}, y_i)
\end{align}
Similarly, the part-level mask loss is given  as:
\begin{align}
\label{eq:Lpm}
   L^{\textrm{pm}}(\hat{m},m) = \frac{1}{N_x C^{\text{part}}} \sum_{0<i<N_x} \sum_{k=1}^{C^{\text{part}}} \ell_M\left(\hat{m}_{\tau_i,k}, m_{i,k} \right)
\end{align}
The total supervised loss for the first stage training of SurgPIS is defined as a weighted sum of instrument-level and part-level losses: $L^{\textrm{sup}} = L^{\textrm{ic}} +L^{\textrm{im}} + L^{\textrm{pm}}$. 
For clarity, we only introduce weighting factors for different losses in 
\secref{sec:impdetails}.

\subsection{Weak PIS Supervision from Disjoint PSS and ISS Datasets}
\begin{figure*}[htb]
  \centering
  \includegraphics[width=\linewidth]{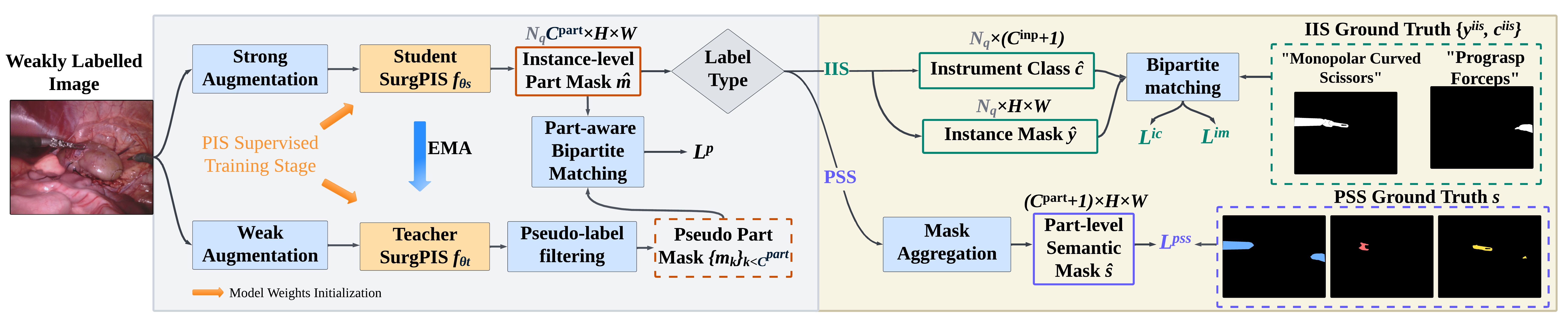}
  \caption{\textit{Weak supervision training for SurgPIS} contains student and teacher SurgPIS models: The teacher SurgPIS generates pseudo-ground-truth \gls{pis} masks for comparison with the student \gls{pis} predictions. The grey part is the shared process, while the green part is the separated process where the student \gls{pis} outputs are aggregated specifically for comparison against either \gls{iis} labels or \gls{pss} labels.\label{fig:ws}}
\end{figure*}

Due to the limited availability of datasets with \gls{pis} labels, we employed a weakly-supervised learning approach to train on datasets containing only \gls{pss} or \gls{iis} labels. Weak supervision for SurgPIS, illustrated in \figref{fig:ws}, consists of a teacher-student model. If ground-truth \gls{pis} labelled training data is available, the student model's predictions are directly compared to it, and the teacher model is bypassed.
If only \gls{pss} or \gls{iis} labelled training data is available, the prediction from the student 
undergoes mask aggregation (cf. \secref{sec:semaggr}) to obtain respectively \gls{pss} or \gls{iis} predictions and compare those to the partial ground truth.

To avoid so-called \emph{catastrophic forgetting} in the student model, which can occur when relying solely on weak supervision with partially labelled ground truth, we supplement its training with the pseudo \gls{pis} supervision from the teacher's prediction.
Both teacher and student process the same input, yet with different augmentations. The input for the teacher model applies weak augmentations (e.g., random flipping), while the student input undergoes strong augmentations, including colour jitter, grayscale conversion, Gaussian blur, and patch erasing.
During training, the teacher model generates pseudo-ground-truth \gls{pis} labels, comprising instrument class probability and part-aware instance masks. 

\subsubsection{Pseudo-ground-truth Filtering and Student-Teacher Updates}
\label{sec:stuteach}
To initialize the student and teacher models for weakly-supervised training, we use the model weights from SurgPIS trained in the first stage.

The teacher model is then updated using the exponential moving average (EMA) of the student weights during training. Let $\theta_{\textrm{stu}}$ and $\theta_{\textrm{teach}}$ be the student and teacher weights at iteration $t$. The teacher update rule is defined as:
\begin{equation}
    \theta_{\textrm{teach}}^{t} = \alpha \theta_{\textrm{teach}}^{t-1} + (1 - \alpha) \theta_{\textrm{stu}}^{t}
\end{equation}
where $\alpha$ is the decay rate that regulates the influence of the student's weights in each iteration. 
This update stabilises training and prevents collapse to trivial solutions.

Inspired by~\cite{filipiak2024polite}, we apply a Dice-based mask scoring filter to retain only pseudo-ground-truth \gls{pis} masks output from the teacher model that have high-confidence masks: \( \text{Dice}(\hat{m}^{\textrm{teach}}_{\tau_i,k},m_{i,k})_ {i\neq 0}>\textrm{Tresh}_{\text{Dice}} \).
Uncertain masks thus contribute zero gradients which stabilizes training~\cite{tarvainen2017mean}.

\subsubsection{Part Semantic Mask Aggregation}
\label{sec:semaggr}
To train SurgPIS on data with only \gls{pss} labels (i.e., $\mathcal{D}_{\textrm{PSS}}$),
we aggregate the part-level mask predictions $\hat{m}_{j,k}$ from the student model into part semantic maps $\hat{s}$.

Generating semantic probability maps for part-only classes is not straightforward, as each part prediction is conditioned on the model's instrument-class probability outputs. To address this dependency, we first construct soft semantic part-level pseudo-probability maps $\rho_k$ based on the model’s predicted instrument-class probabilities.
Each $\hat{y}_j$ and $\hat{m}_{j,k}$ is intrinsically associated with an instrument-class probability vector $\hat{c}_{j}$.
To construct the soft part-level semantic map for the background,
we utilize all instrument masks $\hat{y}_j$ and their associated probability $\hat{c}_j[0]$ corresponding to the background:
\begin{align}
    \rho_0 = \sum_{j} \hat{c}_j[0] \cdot \hat{y}_j
\end{align}
For non-background part classes, we rely on the most probable instrument class:
$\hat{\gamma}_j = \arg\max_{a \leq C^{\textrm{inp}}} \hat{c}_{j}[a]$.
Accumulation of instrument masks, excluding those associated with the background (i.e. $\hat{\gamma}_j>0$), weighted by their associated foreground probability, gives rise to $\rho_k$ for $k>0$:
\begin{align}
    \rho_{k} &= \sum_{j | \hat{\gamma}_j>0} 
    \hat{c}_j[\hat{\gamma}_j] \cdot \hat{m}_{j,k} \, \in \, \mathbb{R}^{H \times W}
\end{align}
We note that 
$[\rho_{0}, \ldots, \rho_{C^{\textrm{part}}}]$ 
is not guaranteed to be composed of valid probability vectors.
To ensure a valid probability distribution at each pixel, we finally apply 
pixel-wise renormalization:
\begin{align}
    \hat{s}[k] &= \frac{ \rho_k }{ \sum_l \rho_l } \, \in \, \Delta^{C^{\textrm{part}} +1}
\end{align}

\subsubsection{Weak Supervision Loss}
To train SurgPIS with weak supervision, we combine the student-teacher \gls{pis} consistency loss $L^{\textrm{sup}}_{\textrm{teach}}$ with a loss on partial ground-truth labels. The consistency loss is computed analogously to the supervised loss $L^{\textrm{sup}}$ from \secref{sec:suloss}, treating the teacher output (\secref{sec:stuteach}) as pseudo-ground-truth and the student output as prediction.
%
For training images from $\mathcal{D}_{\textrm{PSS}}$, we compute the loss between our constructed part-level semantic map $\hat{s}$, obtained in \secref{sec:semaggr}, and the ground truth $s$ via one-to-one  semantic class matching, which can be defined as below:
\begin{align}
    L^{\textrm{wks}}_{\textrm{pss}} =
    \frac{1}{C^{\textrm{part}}} \sum_{k=1}^{C^{\textrm{part}}} \ell_M(\hat{s}[k], s[k])
\end{align}
For training images from $\mathcal{D}_{\textrm{IIS}}$, we calculate the loss between the student instrument-level output and the ground truth \( y^{\text{iis}} \) as discussed in \secref{sec:suloss}:
\begin{align}
L^{\textrm{wks}}_{\textrm{iis}} =
    L^{\textrm{ic}}(\hat{c},c^\text{iis}) + 
    L^{\textrm{im}}
    (\hat{y},y^\text{iis})
\end{align}
Finally, we define the total loss for our weakly-supervised training stage as a sum of the student-teacher consistency loss  and the weak supervision losses (with potential weight factors introduced in \secref{sec:impdetails}):
\begin{align}
    L^{\textrm{wks}} = L^{\textrm{sup}}_{\textrm{teach}} + 
    \begin{cases} 
    L^{\textrm{wks}}_{\textrm{pss}} , ~~~\text{if \gls{pss} label} \\
    L^{\textrm{wks}}_{\textrm{iis}}, ~~~\text{if \gls{iis} label}
    \end{cases}
\end{align}

\section{Experiments}
\label{sec:experiments}

\begin{figure*}[htbp]
\centering
\includegraphics[width=0.85\linewidth]{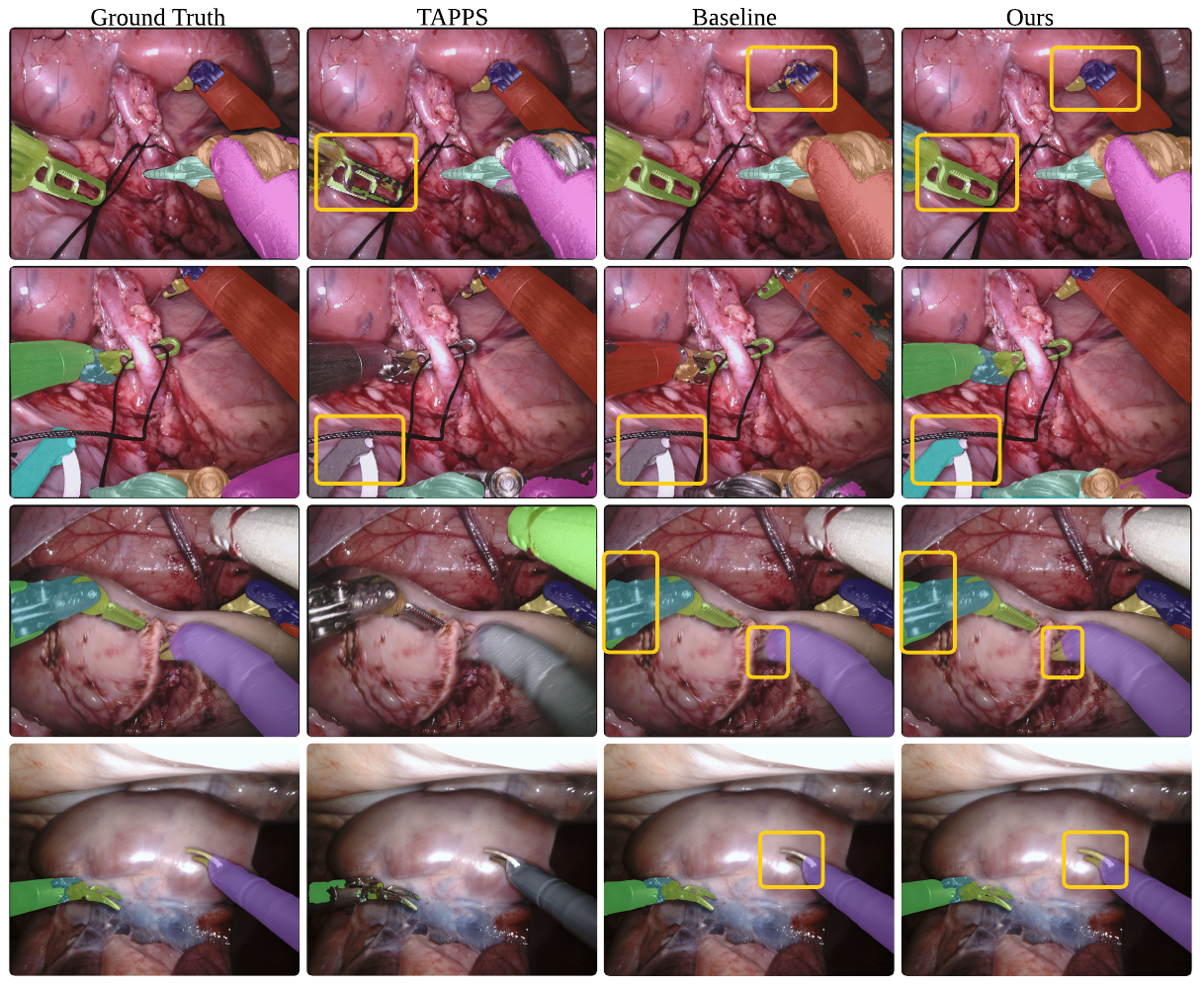}
\caption{Visualization comparing our SurgPIS model with our proposed strong baseline (BPSS$\oplus$BIIS), TAPPS \cite{de2024task}, and the ground truth on the EndoVis2018 testing dataset. Yellow boxes indicate regions of interest where we encourage readers to zoom in, e.g., small instrument parts where SurgPIS has superior performance compared to others.\label{fig:vis}}
\end{figure*}

\begin{table*}[htbp]
\caption{Comparison between SurgPIS and other methods across EndoVis2018 (EV18), EndoVis2017 (EV17), and SAR-RARP50 (SR50) testing set.
Methods in the first and second sections are \gls{iis} and \gls{pss} approaches that are state-of-the-art in the surgical instrument segmentation literature.
Methods in the third section are state-of-the-art \gls{pis} models in the computer vision literature, and our proposed PIS strong baseline (BPSS$\oplus$BIIS). The fourth section is the weak supervision ablation of the disjoint dataset.
\textcolor{black}{\gls{pis}} performance is measured through \textcolor{black}{PartPQ}; \textcolor{black}{\gls{iis}} with \textcolor{black}{PQ} and \textcolor{black}{mc\_IoU}; \textcolor{black}{\gls{pss}} with \textcolor{black}{PartIoU}.}
\centering
\resizebox{\linewidth}{!}{
\begin{tabular}{l|c|ccc|c|cccc|cccc|c}
\hline
\multirow{2}{*}{Method} 
&\multirow{2}{*}{Backbone} 
&\multicolumn{3}{c|}{Training Data\dag} 
& \multirow{2}{*}{Mode*}
&\multicolumn{4}{c|}{EV18} & \multicolumn{4}{c|}{EV17} & SR50 \\ \cline{3-5}\cline{7-15}
& & EV18 & EV17 & SR50 && \textcolor{black}{PartPQ} & \textcolor{black}{PartIoU} & \textcolor{black}{PQ} & \textcolor{black}{mc\_IoU} & \textcolor{black}{PartPQ} & \textcolor{black}{PartIoU} & \textcolor{black}{PQ} & \textcolor{black}{mc\_IoU} & \textcolor{black}{PartIoU} \\
\hline
\multicolumn{15}{l}{\textit{Models trained for \textcolor{black}{ instrument-level instance segmentation (\gls{iis})} task}} \\ [3pt]
\hline
ISINet \cite{gonzalez2020isinet}  & RN-50 & \RIGHTcircle & \RIGHTcircle & \Circle & \sepsymb &- & - & 67.79 & 40.21 & - & -  & 43.85  & 28.96 & -      \\ 
S3Net\cite{baby2023forks} & RN-50 & \RIGHTcircle & \RIGHTcircle & \Circle & \sepsymb & - & - & 69.28 & 42.58 & - & -  & 61.24 & 46.55 & -   \\ 
\hline
\multicolumn{15}{l}{\textit{Models trained for \textcolor{black}{partlevel semantic segmentation (\gls{pss})} task}} \\ [3pt]
\hline
ISINet \cite{gonzalez2020isinet} & RN-50 &\LEFTcircle & \LEFTcircle & \LEFTcircle & \sepsymb & - & 70.24 & - & -& - & 50.73  & -  & -   & 72.91     \\ 
S3Net\cite{baby2023forks}& RN-50 & \LEFTcircle & \LEFTcircle & \LEFTcircle & \sepsymb & - & 71.24 & - & -  & - & 63.83  & - & - & 74.37   \\ 
MATIS \cite{ayobi2023matis} & Swin-S & \LEFTcircle & \LEFTcircle & \LEFTcircle & \sepsymb &- & 78.63 & - & -  & - & 65.37  & - & - & 80.45  \\ 
\hline
\multicolumn{15}{l}{\textit{Models trained for \textcolor{black}{part-aware instance segmentation (\gls{pis})} task}} \\ [3pt]
\hline
PartFormer\cite{li2022panoptic}& RN-50 & \CIRCLE & \CIRCLE & \Circle & \sepsymb & 18.46  & 22.98 & 79.58 & 64.19  &  12.60  & 17.11  & \textbf{73.43} & 46.73 & -\\
PartFormer++\cite{li2024panoptic}& RN-50  & \CIRCLE & \CIRCLE & \Circle & \sepsymb & 21.25  & 24.81 & 82.32 & 63.57  &  13.28 & 18.05  & 73.16 & 49.08 & -\\
TAPPS\cite{de2024task}        &  RN-50    & \CIRCLE & \CIRCLE & \Circle & \sepsymb & 29.85  & 27.65 & 81.65 & 68.11  & 16.72 & 18.84  & 72.30 & 48.12 & -\\
\hline
BPSS$\oplus$BIIS & RN-50 & \RIGHTcircle\LEFTcircle & \RIGHTcircle\LEFTcircle & \Circle & \sepsymb & 61.89  & 78.12 & 72.16 & 67.64  &  52.83  & 68.65  & 63.27 & 48.21 & -\\
BPSS$\oplus$BIIS & RN-50 & \RIGHTcircle\LEFTcircle & \RIGHTcircle & \LEFTcircle & \tgtsymb &63.26  & 80.29 & 78.54 & 67.02  & 53.07 & 70.80  & 66.98  & 49.37 & 78.72  \\ 
BPSS$\oplus$BIIS & Swin-B & \RIGHTcircle\LEFTcircle & \RIGHTcircle & \LEFTcircle & \tgtsymb &63.43  & 79.27 & 78.99 & 68.41  & 55.49 &  69.26  &  68.37 & 50.76  & 79.62  \\
\hline
SurgPIS (ours) & RN-50 & \CIRCLE & \CIRCLE & \Circle &  \sepsymb &72.96  & 80.54 & 81.03 & 67.47 & 64.06 &70.28 & 71.43 & 46.09 & -\\
SurgPIS (ours) & RN-50 & \CIRCLE & \RIGHTcircle & \LEFTcircle & \tgtsymb & 75.73  & 83.54 & \textbf{84.07} & \textbf{72.85} & 68.50 & 75.76 & 71.85 & 49.13 & 83.69 \\
SurgPIS (ours) & Swin-B & \CIRCLE & \RIGHTcircle & \LEFTcircle &  \tgtsymb & \textbf{77.92}  & \textbf{85.66} &  83.31 & 70.31 & \textbf{69.14} & \textbf{77.02} & 72.94 & \textbf{52.81}  & \textbf{84.75} \\
\hline
\multicolumn{15}{l}{\textit{Ablation on  weak \textcolor{black}{\gls{pis}} supervision only from \textcolor{black}{\gls{pss}} or \textcolor{black}{\gls{iis}} dataset}} \\ [3pt]
\hline
SurgPIS (ours) & RN-50  & \CIRCLE & \RIGHTcircle & \Circle & \tgtsymb & 73.43  & 80.95 & 82.58 & 69.82 & 65.66 & 72.94 & 70.67 & 45.33  & -\\
SurgPIS (ours) & RN-50  & \CIRCLE & \Circle & \LEFTcircle& \tgtsymb & 72.21  & 81.16 & 82.73 & 69.98 & \iffalse 54.31 & 62.95 & 63.34 & 64.33 \else - & - & - & - \fi  & 80.26\\
\hline
\end{tabular}}
\begin{threeparttable}
        \begin{tablenotes}
        \small
            \item \dag Training Data -- 
            \CIRCLE{}: Data with PIS labels.
            \LEFTcircle{}: Data with PSS labels. 
            \RIGHTcircle{}: Data with IIS labels.
            \Circle{}: Data not used
             \item *Mode -- \tgtsymb{}: Train on combined datasets. \sepsymb{}: Train on a single dataset corresponding to the testing sets
        \end{tablenotes}
 \end{threeparttable}
\label{tab:pis_comparison}
\end{table*}

\subsection{Datasets}
We train and evaluate our method on four public surgical instrument segmentation datasets for robotic surgery.
\begin{enumerate}
\item 
\emph{EndoVis2017}~\cite{allan20192017} comprises 10 videos captured using the da Vinci robotic system with 7 distinct surgical instrument types: bipolar forceps (BF), prograsp forceps (PF), large needle driver (LND), vessel sealer (VS), grasping retractor (GR), monopolar curved scissors (MCS) and ultrasonic probe (UP).
We manually created \gls{iis} annotations using the existing \gls{iss} label for all training and testing frames by assigning a unique instance ID to each instrument occurrence, maintained consistently across temporally adjacent frames. The new \gls{iis} annotations are combined with the existing \gls{pss} ones to construct \gls{pis} labels, where each PIS mask represents \textit{(Instrument type -- Instance ID -- Part type)}.

\item 
\emph{EndoVis2018}~\cite{allan20202018} includes 15 video sequences (11 for training, 4 for testing), covering 7 predefined instrument categories: bipolar forceps, prograsp forceps, large needle driver, monopolar curved scissors, ultrasound probe, suction instrument (SI), and clip applier (CA).
It was additionally annotated by \cite{baby2023forks,gonzalez2020isinet} for \gls{iis} labels, which we leveraged to construct \gls{pis} annotations by combining with existing \gls{pss} labels.

\item 
\emph{SAR-RARP50}~\cite{psychogyios2023sar} contains 50 suturing video segments of robotic-assisted radical prostatectomy. 
It provides only semantic part labels, which we used for \gls{pss} task evaluation.  
The official test sets of all three datasets were used for testing purposes.

\item
To the best of our knowledge, there are currently no publicly available and validated datasets that provide the appropriate scale of annotations for our \gls{pis} task.
For further cross-dataset generalisation evaluation, we use GraSP~\cite{ayobi2024pixel}, which 
provides \gls{iis} annotations, and we hold it unseen during training. We manually annotate the \gls{pis} labels for all testing images in the \emph{GraSP}~\cite{ayobi2024pixel} dataset using an interactive labelling tool~\cite{he2024segment} built upon SAM~\cite{kirillov2023segment}. All GraSP \gls{pis} annotations were manually verified and corrected by human experts with knowledge of surgical instrument anatomy. SAM served only as an interactive tool to accelerate the annotation process, with final labels reflecting human expert verification. Note that all instrument classes in the GraSP dataset are covered by those in EndoVis2017 and EndoVis2018. 
We only re-categorised the original laparoscopic grasper in GraSP as the grasping retractor (GR), as they refer to the same instrument. 
\end{enumerate}
Across all \gls{pis}-annotated datasets, we adopt a unified part taxonomy comprising $C^{\text{part}}$ part classes in a fixed, predefined order (e.g., shaft, wrist, clasper). This ordering is critical for our 
part-specific query transformation module (\secref{sec:pisgen}), which explicitly maps each instrument-level query to its corresponding ordered part queries.
\subsection{Implementation Details}
\label{sec:impdetails}
For all datasets, we use a
batch size of 16 images, and train on 2 Nvidia A100 GPUs. We use the AdamW optimizer following~\cite{cheng2022masked} with a weight decay
of 0.05, and a polynomial learning rate decay schedule with an initial learning rate of $10^{-4}$
and a power of 0.9. 
In the first stage of training, we train SurgPIS for 40k iterations and apply large scale jittering with a scale between 0.1 and 2.0, followed by a random crop of 512×512 pixels for augmentation. 
The weighting factors for instrument-level classification loss $\alpha_{class}$ and mask loss $\alpha_{mask}$ are both set to 5.0, with the mask loss covering both instrument-level and part-level mask losses. In the second stage of training, we set the EMA decay rate to $\alpha = 0.99$, and the weighting factor for loss with \gls{pss} labeled data and \gls{iis} labelled data are 3.0 and 2.0, respectively.
Our source code and annotations will be made open access upon publication.

\subsection{Proposed Strong Baseline (BPSS\texorpdfstring{$\oplus$}{+}BIIS)}
As this is the first work on \gls{pis} in the surgical domain,  in addition to our proposed SurgPIS, we also establish a strong baseline using Mask2Former~\cite{cheng2022masked} with RN-50~\cite{he2016deep} and Swin-B~\cite{liu2021swin} backbones by following the approach of~\cite{kirillov2019panoptic}. 
We first train separate baseline models for \gls{pss} (B\gls{pss}) and \gls{iis} (B\gls{iis}). 
To construct our \gls{pis} baseline (BPSS$\oplus$BIIS), we simply obtain each binary instrument instance mask from B\gls{iis} and multiply it with all binary part segmentation masks from B\gls{pss}. $B_{PSS}\oplus B_{IIS}$  implements two-stage hierarchical approach which makes no use of end-to-end part-instance association.

\subsection{Evaluation Metrics}
For \gls{pis} task, we use the part-aware panoptic quality (PartPQ) metric~\cite{de2021part,de2024task}, which relies on the part-level mean IoU within a pair of matched instance ($\textrm{IOU}_\text{p}$). 
The PartPQ for the testing set with instrument-level classes $C$ is defined as follows:
\begin{equation}
\text{PartPQ} = \frac{1}{C}\sum^{C}_{c=1}\frac{\sum_{(\hat{m}_{\tau_i}, m_i) \in TP_c} \text{IOU}_\text{p}(\hat{m}_{\tau_i}, m_i)}{|TP_c| + \frac{1}{2} |FP_c| + \frac{1}{2} |FN_c|}
\end{equation}  
where \( TP_c \), \( FP_c \), and \( FN_c \) denote the sets of true positive, false positive, and false negative segments, respectively. 
A matched instances index pair $(i,\tau_i)$
is included in \( TP_c \) if their instrument-level masks have an $\textrm{IOU}_\text{p}$ greater than 0.5. 
Unidentified ground-truth instances belong to \( FN_c \), while incorrect predictions are classified as \( FP_c \). 

We evaluate the PSS task by mean Intersection-over-Union (IoU) across the part types across all the frames in the testing set ($\textrm{IOU}_\text{k}$), denoted as PartIoU:%
\begin{equation}
\text{PartIoU} =\frac{1}{K}\sum_{k = 1}^{K}\textrm{IOU}_\text{k}(\hat{s}[k], s[k])
\end{equation}
where $\hat{s}$ is our constructed part-level semantic map, as shown in \secref{sec:semaggr}, and $s$ is ground truth.

To evaluate the multi-class instrument segmentation (\gls{iss} task), following the evaluation method from \cite{gonzalez2020isinet}, we employ three  IoU-based metrics commonly used in the surgical domain~\cite{gonzalez2020isinet}: mc\_IoU, ISI\_IoU, and mc\_IoU. 
For each instrument-level class $c$ with total classes $C$, Ch\_IoU computes IoU only over frames where the class $c$ appears in the \gls{iis} ground truth $y^{\text{iis}}$ as:
\begin{equation}
\text{Ch\_IoU} = \frac{1}{\mathcal{F}_c} \sum_{f = 1}^{\mathcal{F}_c} \big(\frac{1}{C}\sum_{c = 1}^{C}
\text{IOU}(\hat{y}_{c,f}, y_{c,f}^{\text{iis}})\big)
\end{equation}
where \( \mathcal{F}_c = \{ f \mid \text{instrument } c \text{ present in } y^{\text{iis},(f)} \} \), which leads to excluding the model's false positive prediction. The ISI\_IoU is a variation of Ch\_IoU averaged over
all the frames $\mathcal{F}$:
\begin{equation}
\text{ISI\_IoU} = \frac{1}{\mathcal{F}} \sum_{f = 1}^{\mathcal{F}} \big(\frac{1}{C}\sum_{c = 1}^{C}
\text{IOU}(\hat{y}_{c,f}, y_{c,f}^{\text{iis}})\big)
\end{equation}
In contrast, mc\_IoU calculates the instrument-level class mean IoU ($\textrm{IOU}_\text{c}$) across all frames globally, defined as:
\begin{equation}
\text{mc\_IoU} = \frac{1}{C}\sum_{c = 1}^{C} 
\textrm{IOU}_\text{c}(\hat{y}_{c}, y_{c}^{\text{iis}})\big)
\end{equation}

The mIoU for various instrument classes is reported under the acronyms of the instrument names.

\subsection{Results} 
\subsubsection{Comparison to the State-of-the-Art in PIS}
\tabref{tab:pis_comparison} shows our model surpasses the strong \gls{pis} baseline (BPSS$\oplus$BIIS) by 11.07 percentage points (pp) in PartPQ when trained on EndoVis2018 in a fully-supervised setting. Similarly, when trained only on EndoVis2017, our model achieves 11.23~pp improvement in PartPQ. Unlike the baseline, which requires training separate models for \gls{pss} and \gls{iis} to obtain \gls{pis} results, our unified approach enables part features to be directly informed by instrument-level features, enhancing the performance for \gls{pis} task while simplifying the training pipeline. 

Existing part-aware panoptic models, 
TAPSS~\cite{de2021part} and PartFormer(++)~\cite{li2022panoptic,li2024panoptic},
originally designed for natural images, 
excel in \gls{iis} task (PQ, mc\_IoU) but underperform in \gls{pis} (PartPQ).
In contrast, our method significantly improves PartPQ over TAPPS \cite{de2024task}, showcasing SurgPIS superiority in the surgical domain. Notably, while TAPPS \cite{de2024task} also adopts a shared-query approach, our SurgPIS achieves substantially higher performance on the PartPQ score when training separately for EndoVis2017 and EndoVis2018, highlighting the effectiveness of our part-specific query transformation.

\begin{figure}[htbp]
  \centering
  \includegraphics[width=\linewidth]{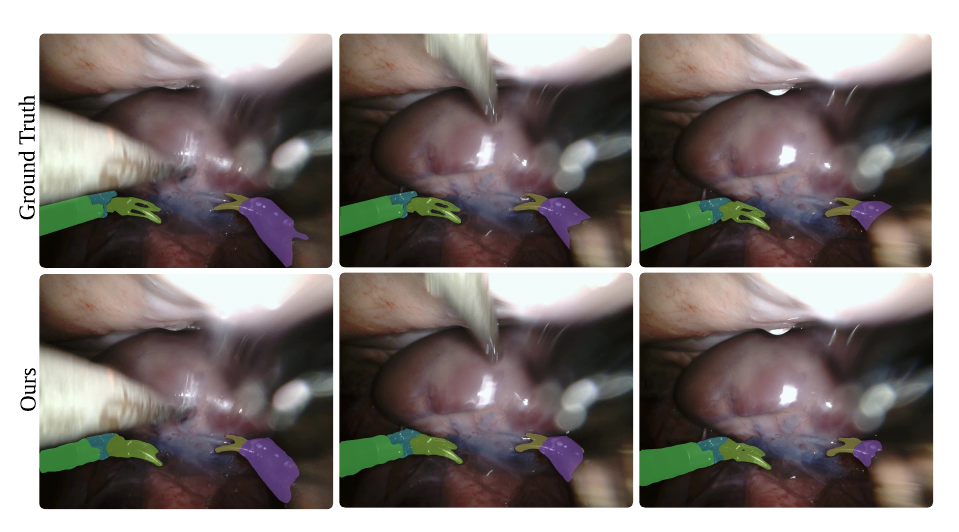}
  \caption{challenging examples of consecutive frames with severe motion blur and specular reflection from the EndoVis2018 dataset.\label{fig:motion_blur}}
\end{figure}

When training on all datasets, our model outperforms the baseline by 14.49 pp in PartPQ under a weakly supervised setting on EndoVis2018 and by 13.65 pp on EndoVis2017, both using the Swin-B backbone (same observation when using RN-50 backbone), highlighting SurgPIS's ability to leverage diverse label types, effectively addressing the scarcity of \gls{pis} annotations in surgical instrument datasets. By integrating different granular labels during training, our model learns richer feature representations, leading to improved part-level segmentation performance. \figref{fig:vis} further illustrates SurgPIS's qualitative advantages. Notably, on challenging consecutive frames from the EndoVis2018 dataset exhibiting extreme motion blur and strong specular reflections, as shown in \figref{fig:motion_blur}, SurgPIS maintains consistent and accurate segmentation performance for part-level and instrument-level segmentation compared to the ground truth.

\begin{figure}[htbp]
  \centering
\includegraphics[width=\linewidth]{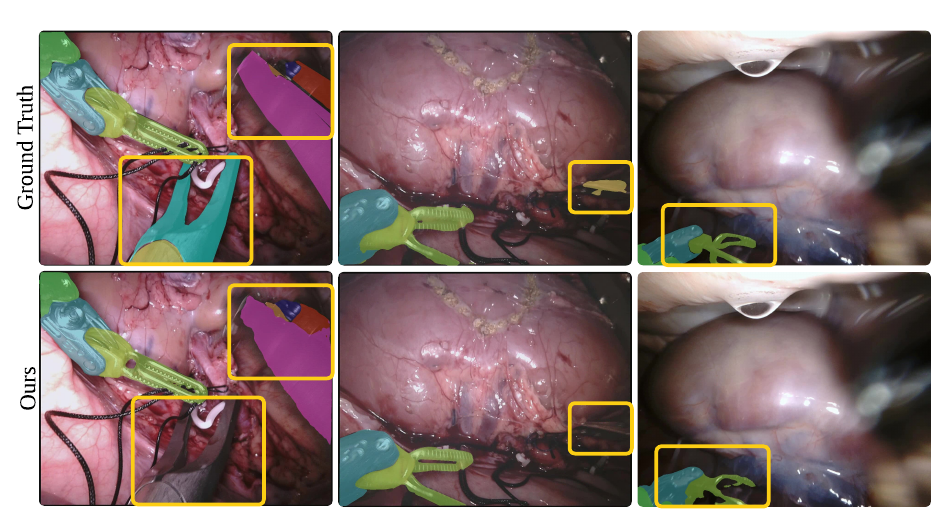}
  \caption{Representative failure cases from EndoVis2018 demonstrating performance degradation under challenging conditions. Yellow boxes highlight regions of interest where our model exhibits reduced accuracy.}\label{fig:f_Case}
\end{figure}

We present representative failure cases in \figref{fig:f_Case}. Our model exhibits limitations under challenging visual conditions. High-speed instrument motion induces severe motion blur, which degrades the instruments' edge sharpness and distorts part geometry. Highly reflective tissue surfaces produce ambiguous appearance cues that the model misinterprets as instrument parts. These effects lead to missed detections, false positives, and imprecise boundary segmentation.

\begin{table*}[tbh]
\caption{Comparison of semantic segmentation performance (\textcolor{black}{\gls{iss}} task) between our SurgPIS (trained for \textcolor{black}{\gls{pis}}) and state-of-the-art models trained for either \textcolor{black}{\gls{iis}} or \textcolor{black}{\gls{iss}}.
Results are reported separately for EndoVis2017 (EV17) and EndoVis2018 (EV18).\label{tab:iss_comparison}}
\centering
\begin{subtable}{\linewidth}
    \centering
    \resizebox{\linewidth}{!}{
    \begin{tabular}{l|c|c|c|c|ccccccc|c}
    \hline
        Method &  Task & Training Data\dag: EV17& Ch\_IoU & ISI\_IoU & BF & PF & LND & VS & GR & MCS & UP & mc\_IoU \\ \hline
        ISINet \cite{gonzalez2020isinet} & \textcolor{black}{\gls{iis}}& \RIGHTcircle & 55.62 & 52.20 & 38.70 & 38.50 & 50.09 & 27.43 & 2.01 & 28.72 & 12.56 & 28.96 \\
        TraSeTR \cite{zhao2022trasetr}& \textcolor{black}{\gls{iis}}& \RIGHTcircle & 60.40 & 65.20 & 45.20 & 56.70 & 55.80 & \textbf{38.90} & 11.40 & 31.3 & 18.20 & 36.79 \\
        S3Net \cite{baby2023forks} & \textcolor{black}{\gls{iis}}& \RIGHTcircle & 72.54 & \textbf{71.99} & 75.08 & 54.32 & 61.84 & 35.5 & 27.47 & \textbf{43.23} & 28.38 & \textbf{46.55} \\ 
        MF-TAPNet \cite{jin2019incorporating} &\textcolor{black}{\gls{iss}}& $\diamondrightblack$ & 37.35 & 13.49 & 16.39 & 14.11 & 19.01 & 8.11 & 0.31 & 4.09 & 13.40 & 10.77 \\
        MATIS \cite{ayobi2023matis} & \textcolor{black}{\gls{iss}} &$\diamondrightblack$ & 71.36 & 66.28 & 68.37 & 53.26 & 53.55 & 31.89 & 27.34 & 21.34 & 26.53 & 41.09 \\ \hline
        SurgPIS (ours) (RN-50) & \textcolor{black}{\gls{pis}}& \CIRCLE & \textbf{75.92} & 69.81 & \textbf{79.13} & \textbf{58.11} & \textbf{63.40} & 20.19 & \textbf{29.68} & 41.94 & \textbf{30.15} & 46.09 \\\hline
    \end{tabular}
    }
\end{subtable}
    \begin{subtable}{\linewidth}
        \centering
        \resizebox{\linewidth}{!}{
            \begin{tabular}{l|c|c|c|c|ccccccc|c}
            \hline
                Method & Task & Training Data\dag: EV18 & Ch\_IoU & ISI\_IoU & BF & PF & LND & SI & CA & MCS & UP & mc\_IoU \\ \hline
                ISINet \cite{gonzalez2020isinet} & \textcolor{black}{\gls{iis}}& \RIGHTcircle & 73.03 & 70.97 & 73.83 & 48.61 & 30.98 & 37.68 & 0.00 & 88.16 & 2.16 & 40.21 \\
                TraSeTR \cite{zhao2022trasetr} & \textcolor{black}{\gls{iis}}& \RIGHTcircle & 76.20  & - & 76.30 & 53.30 & 46.50 & 40.60 & \textbf{13.90} & 86.30 & 17.50 & 47.77 \\
                S3Net \cite{baby2023forks} & \textcolor{black}{\gls{iis}}& \RIGHTcircle & 75.81 & 74.02 & 77.22 & 50.87 & 19.83 & 50.59 & 0.00 & 92.12 & 7.44 & 42.58 \\
                MF-TAPNet \cite{jin2019incorporating} &\textcolor{black}{\gls{iss}} & $\diamondrightblack$ & 67.87 & 39.14 & 69.23 & 6.10 & 11.68 & 14.00 & 0.91 & 70.24 & 0.57 & 24.68 \\
                MATIS \cite{ayobi2023matis} &\textcolor{black}{\gls{iss}}& $\diamondrightblack$ & 84.26 & 79.12 & 83.52 & 41.90 & 66.18 & 70.57 & 0.00 & \textbf{92.96} & 23.13 & 54.04 \\ \hline
                SurgPIS (ours) (RN-50) & \textcolor{black}{\gls{pis}}& \CIRCLE & \textbf{87.86} & \textbf{83.75} & \textbf{87.98} & \textbf{82.69} & \textbf{80.08} & \textbf{77.94} & 5.12 & 92.83 & \textbf{45.65} & \textbf{67.47} \\\hline
            \end{tabular}
        }
    \end{subtable}
    \begin{threeparttable}
        \begin{tablenotes}
        \small
            \item \dag Training Data -- \CIRCLE{}: Data with \gls{pis} labels.
            $\diamondrightblack$: Data with \gls{iss} labels.
            \RIGHTcircle{}: Data with \gls{iis} labels.
        \end{tablenotes}
    \end{threeparttable}
\end{table*}
\addtocounter{table}{-1}


\subsubsection{Comparison to the State-of-the-Art in IIS and PSS}
We evaluate our model on \gls{iis} and \gls{pss} tasks by aggregating the \gls{pis} predictions into \gls{iis} and \gls{pss} ones and comparing these to existing surgical instrument segmentation models as shown in \tabref{tab:pis_comparison}. 
When solely trained on EndoVis2018, our model surpasses ISINet \cite{gonzalez2020isinet} trained for \gls{iis} task by 13.24 pp in PQ, suggesting training with PIS labels can improve the model instrument-level instance understanding. 
Our model, trained in a weakly-supervised setting on the SAR-RARP50 dataset, outperforms MATIS \cite{ayobi2023matis} and the strong baseline, which are trained specifically for the \gls{pss} task in a fully-supervised setting, by 4.3 pp and 5.13 pp in PartIoU, respectively. 
Unlike these single-task models, our approach not only enables \gls{pis}, a capability they lack, but also achieves strong performance in both \gls{pss} and \gls{iis} tasks, offering a more unified solution.

\subsubsection{Comparison to the State-of-the-Art in ISS}
We evaluate the performance of SurgPIS and other state-of-the-art surgical instrument segmentation models on the \gls{iss} task in \tabref{tab:iss_comparison}. 
The existing models are trained either for \gls{iss} or \gls{iis} task (noting that \gls{iis} is richer than \gls{iss} and can be aggregated to it) using corresponding label types. 
Despite being designed for \gls{pis}, SurgPIS outperforms the best-performing competing model, S3Net~\cite{baby2023forks}, with a 3.38 pp gain in Ch\_IoU on EndoVis2017 and a 3.6 pp improvement on EndoVis2018 comparing to the state-of-the-art model MATIS~\cite{ayobi2023matis}. 
It also achieves the highest ISI\_IoU scores across both datasets while maintaining strong performance across different instrument classes. Notably, all methods use the same training data, but SurgPIS is trained with \gls{pis} labels, whereas others rely on \gls{iss} or \gls{iis} labels. 
These results highlight the benefits of reformulating surgical instrument segmentation as a \gls{pis} task, enabling simultaneous multi-granularity segmentation while improving instrument-level performance through 
finer-grained annotations.
We further compare with the recent SurgSAM2~\cite{liu2024surgical}, which reports
84.4 Ch\_IoU on EndoVis2018 under full-mask prompting, compared with
87.86 achieved by SurgPIS without inference-time prompting.
We note that this comparison is not strictly direct, since SurgSAM2 performs
prompt-driven video object segmentation, whereas SurgPIS automatically performs
the more comprehensive \gls{pis} task.

\subsubsection{Generalisation Evaluation on Unseen Data}
\begin{figure}[htbp]
\centerline{\includegraphics[width=\columnwidth]{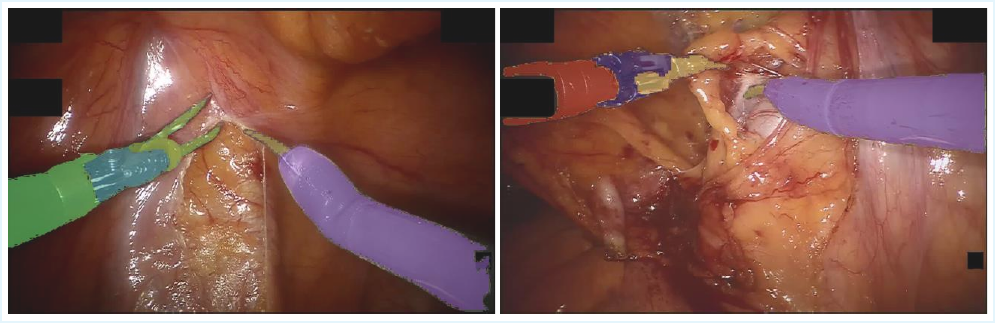}}
\caption{Visualization examples of our SurgPIS model trained on EndoVis2017, EndoVis2018, and SAR-RARP50 but evaluated on the GraSP testing dataset.\label{fig:vis_grasp}}
\end{figure}

We assess the generalisation performance of our SurgPIS model pre-trained on the combined EndoVis2017, EndoVis2018, and SAR-RARP50 by testing it on the GrasP testing set annotated for \gls{pis}. 
We compare with other pre-trained state-of-the-art models.
\tabref{table:grasp} demonstrates the 
generalisation capability
of SurgPIS on the external GraSP dataset. Specifically, 
the baseline model TAPIS \cite{ayobi2024pixel}, when trained and tested directly on the GraSP dataset(i.e., same dataset, no generalization), performs 73.42 PQ and  61.12 mc\_IoU. Our PQ dropped by only 5.14 pp and mc\_IoU by 2.39 pp compared to the TAPIS, indicating SurgPIS still achieves competitive generalized performance on the \gls{iis} task.
This highlights the model's cross-dataset generalizability at the instrument level, which is more challenging and clinically relevant than in-dataset evaluation, demonstrating consistent performance across different surgical scenarios for the same instrument classes. For the \gls{pis} task, the ability lacking in the GraSP benchmark, SurgPIS maintained performance consistent with the results shown in \tabref{tab:pis_comparison}, consistently outperforming state-of-the-art models. Notably, it achieved a 32.85 pp improvement in PartPQ over TAPPS \cite{de2024task}, and even surpassed the strong baseline BPSS$\oplus$BIIS by 5.51 pp in PartPQ. These external validation results underscore SurgPIS’s robustness not only for \gls{iis}, but also for the more challenging \gls{pss} and \gls{pis} tasks, 
demonstrating its efficacy on different tasks across varying surgical environments, lighting, and imaging conditions. The visualization examples of the pretrained SurgPIS model testing on the GraSP dataset are shown as~\figref{fig:vis_grasp}. 

\begin{table}[htbp]
\caption{
\gls{pis} models
trained on EndoVis2018, EndoVis2017, and SAR-RARP50
but evaluated on the GraSP testing set.}
\centering
\resizebox{\linewidth}{!}{
\begin{tabular}{l|cccc}
\hline
Model & PartPQ & PartIoU & PQ & mc\_IoU\\ \hline 
PartFormer \cite{li2022panoptic} & 24.83 & 28.56 & 64.87 &  54.96\\
PartFormer++ \cite{li2024panoptic}  & 26.57  & 29.14 & 62.07 & 57.12 \\
TAPPS \cite{de2024task}& 27.83 & 31.09 & 63.93 & 51.28\\
BPSS$\oplus$BIIS & 55.17 & 63.88 & 61.79 & 56.96\\
\hline
SurgPIS & \textbf{60.68} & \textbf{67.15} & \textbf{68.28} & \textbf{58.73}\\
\hline
\end{tabular}
}
\label{table:grasp}
\end{table}
\begin{table*}[tbp]
\centering
\caption{Comparison of SurgPIS's extended part-aware panoptic segmentation model with the nnUNet framework. Ours is trained on \gls{pis} labels of surgical instruments from EndoVis 2017, EndoVis 2018, and SAR-RARP50, and nnU-Net is trained on \gls{pss} labels of surgical instruments; both further incorporate additional instrument and anatomy semantic annotations from EndoVis 2018 and SAR-RARP50.}
\label{tab:panoptic_extension}
\resizebox{\textwidth}{!}{
\begin{tabular}{l|c|cccccccccc|c}
\hline
\multicolumn{13}{c}{Testing on EndoVis2018} \\
\hline
Method & Backbone &Parenchyma & Covered kidney& Shaft & Clasper & Wrist & Thread & Needle & US probe & Intestine & Clips & mc\_IoU\\
\hline
nnUNet \cite{isensee2021nnu} & - &72.22 & 58.21 & 83.14 & 63.04 & 58.82 & 51.09 & 49.28 & 65.46 & 67.68 & 84.38 & 65.33 \\
Ours & RN-50 & 72.38 & \textbf{63.09} & 86.63 & 84.27 & 85.92 & \textbf{65.98} & \textbf{63.51} & 70.76 & 73.12 & \textbf{86.58} & 75.22\\
Ours & Swin-B &\textbf{73.26} & 62.13 & \textbf{87.52} & \textbf{86.93} & \textbf{87.51} & 63.09 & 62.10 & \textbf{72.37} & \textbf{74.85} & 85.96 & \textbf{75.57}\\
\hline
\end{tabular}%
}

\resizebox{\textwidth}{!}{
\begin{tabular}{l|c|ccccccccc|c}
\multicolumn{12}{c}{Testing on SAR-RARP50} \\
\hline
Method & Backbone &Shaft & Clasper & Wrist & Thread & Suturing needle & Suction tool & Needle Holder & Clamps & Catheter & mc\_IoU\\
\hline
nnUNet \cite{isensee2021nnu} & - &71.28 & 62.32 & 58.07 & 52.98 & 53.46 & 69.96 & 70.58 & 40.57 & 56.33 & 59.51  \\
Ours & RN-50 & 83.19 & 80.07 & 80.68 & \textbf{59.12} & \textbf{56.39} & 68.23 & 72.31 & \textbf{43.33} & 59.58 & \textbf{66.99} \\
Ours & Swin-B & \textbf{85.21} & \textbf{81.39} & \textbf{82.56} & 56.32 & 47.27 & \textbf{75.06} & \textbf{74.85} & 35.28 & \textbf{60.39} & 66.48 \\
\hline
\end{tabular}%
}
\end{table*}

\subsubsection{Extensions of Surgical Part-aware Panoptic Segmentation}
To assess SurgPIS's effectiveness to a broader range of other instruments (e.g., suturing needle, thread, clamps) and anatomical structures (e.g., arenchyma, covered kidney, intestine), we extend our model to the task of part-aware panoptic segmentation. The architecture of SurgPIS can be naturally adapted to this task by incorporating a standard semantic segmentation component for "stuff" classes, following Mask2Former \cite{cheng2022masked}. We keep all training settings unchanged as detailed in \secref{sec:method}, but incorporate semantic masks ('stuff' classes) for other instruments and anatomy from both the EndoVis2018 and SAR-RARP50 datasets. During training, masks are supervised accordingly; During inference, for semantic segmentation, predicted segments ($\hat{y}_j$) sharing the same class label ($\hat{c}_j$) are merged, as discussed in \secref{sec:Notations}. For instance segmentation, the unique index $j$ distinguishes different instances of the same instrument class.

For comparison, we train nnUNet~\cite{isensee2021nnu}, a strong baseline in medical imaging with demonstrated capability for segmenting thin structures, as a pure semantic segmentation model where all instances of the same part type (e.g., all shafts across different instruments) are treated as a single semantic class. In contrast, our part-aware panoptic segmentation approach explicitly distinguishes part instances belonging to different instrument instances. SurgPIS predictions for \gls{pis} masks of surgical instruments are aggregated to part-level semantics for fair comparison with nnUNet.

As shown in \tabref{tab:panoptic_extension}, SurgPIS outperforms nnUNet~\cite{isensee2021nnu}. On EndoVis 2018, our model with the Swin-B backbone achieves state-of-the-art mc\_IoU with a substantial +10.24 pp improvement over nnUNet. This gain is particularly pronounced on fine-grained surgical instrument parts, with +23.89 pp mean IoU on Claspers and +28.99 pp on Wrists. These results generalize effectively to SAR-RARP50, where SurgPIS demonstrates significant improvements on surgical parts such as the Wrist (+24.49 pp on mean IoU). Due to patch size downsampling in transformer backbones like Swin-B, our convolutional variant with RN-50 better captures slim structures. On SAR-RARP50, it outperforms nnUNet by +6.17~pp mean IoU on the Thread class, showing that SurgPIS maintains strong performance on small and slim instruments.
Overall, SurgPIS excels at segmenting surgical instrument parts while maintaining robust performance across diverse anatomical contexts and small instruments, demonstrating strong potential for real-world surgical applications.

\subsubsection{Ablation Study}
We conduct ablation studies in \tabref{tab:ablation} for SurgPIS with a ResNet-50 backbone, which is trained and tested on EndoVis2018, EndoVis2017, and SAR-RARP50.

\paragraph{Part-aware Bipartite Matching}
We assess the impact of our part-aware bipartite matching module by replacing it with the original bipartite matching in Mask2Former~\cite{cheng2022masked} (w/o PBM), showing a 7.79 pp drop on EndoVis2018 and 8.56 pp on EndoVis2017 in PartPQ, and 15.3 pp on SAR-RARP50 in PartIoU compared to SurgPIS that utilise part-level loss in bipartite matching, highlighting the importance of integrating part-level information into the matching process. It also suggests that accurate part-level segmentation learned from \gls{pis} data is crucial for generalising to \gls{pss} supervision.

\paragraph{Part-specific Query Transformation}
We replace our part-specific queries (w/o PSQ) with the generic part queries proposed in TAPSS \cite{de2021part}, which leads to a notable decline of 34.94 pp on EndoVis2018 and 39.67 pp on EndoVis2017 in PartPQ, and 23.62 pp in PartIoU on SAR-RARP50. This underscores the importance of designing specific queries for identical part types across different instrument classes.

\paragraph{Weak Supervision Ablation}
For the weakly supervised training stage, we analyse the impact of weak \gls{pis} supervision from disjoint \gls{pss} and \gls{iss} datasets. Removing \gls{pss} supervision (i.e., excluding SAR-RARP50) or \gls{iss} supervision (i.e., excluding EndoVis2017), as shown in the 4th and 3rd last rows of \tabref{tab:pis_comparison}, results in a performance drop across all metrics. This highlights the effectiveness of the benefit of incorporating additional \gls{iis} supervision and \gls{pss} supervision via our part-semantic mask aggregation strategy for improved model performance.
We also analyse that when removing the pseudo-label filtering module (w/o PLF), for \gls{iis} labelled data, the performance drops by 9.74 pp on PartPQ and 11.46 pp on PQ, and for \gls{pss} labelled data, it reduces by 4.73 pp on PartIoU, which suggests that uncertain masks have a greater impact on the calculation for weak supervision loss. 
When freezing the pre-trained teacher model without EMA updates, performance drops by 1.06 pp and 3.14 pp in PartPQ on EndoVis2018 and EndoVis2017, respectively, indicating that continuous teacher improvement through EMA is effective for pseudo-label generation across disjoint datasets.
Applying the same weak augmentation to the input data (w/ Same Aug) leads to an 8.31 pp drop in PQ for \gls{iis} labelled data and a 5.75 pp drop in PartIoU for \gls{pss} labelled data, which suggests that enforcing consistency between different training signals helps the model generalise better on the weakly labelled data.    
\begin{table}[htb!]
    \caption{Ablation study for SurgPIS with a ResNet-50 backbone trained and tested on \gls{pis} labelled dataset EndoVis2018 (EV18), \gls{iis} labelled dataset EndoVis2017 (EV17), and \gls{pss} labelled dataset SAR-RARP50 (SR50).}
    \centering
    \resizebox{\linewidth}{!}{
    \begin{tabular}{l|cc|cc|c}
    \hline
        \multirow{2}{*}{Ablation} & \multicolumn{2}{c|}{EV18}& \multicolumn{2}{c|}{EV17} & SR50 \\  
        \cline{2-6}
        & PartPQ & PQ & PartPQ & PQ & PartIoU\\  
        \hline
        w/o PBM & 67.94 & 82.73 & 59.94  & 70.83 &  68.39\\ 
        w/o PSQ & 40.79 & 81.67 & 28.83 & 71.46 & 60.07 \\ 
        w/o PLF & 75.02 &  84.75 & 58.76 & 60.39 & 78.96\\
        w/o EMA & 4.67 &  82.75 & 65.09 & 69.05 & 78.96\\
        w/ Same Aug & 74.82 & 82.51 & 60.23 & 63.54 & 77.94\\
        \hline
        SurgPIS & 75.73 & 84.07 & 68.50 & 71.85& 83.69\\
    \hline
    \end{tabular}
}
    \label{tab:ablation}
\end{table}
\begin{table*}[htbp]
\caption{Computational efficiency vs. performance trade-offs on EndoVis2017, EndoVis2018, SAR-RARP50 (SR50), and GraSP for different backbones, including pre-trained foundation models (\dag).}
\centering
\resizebox{\linewidth}{!}{
\begin{tabular}{l|ccc|cccc|cccc|c|cccc}
\hline
\multirow{2}{*}{Backbone} & \multirow{2}{*}{Params} & \multirow{2}{*}{FLOPs} & \multirow{2}{*}{FPS} & \multicolumn{4}{c|}{EndoVis2018} & \multicolumn{4}{c|}{EndoVis2017} &SR50 & \multicolumn{4}{c}{GraSP}\\
\cline{5-17}
&&&&PartPQ & PartIoU & PQ & mc\_IoU & PartPQ & PartIoU & PQ & mc\_IoU & PartIoU &  PartPQ & PartIoU & PQ & mc\_IoU\\
\hline
RN-50 & 44.2M & \textbf{61.29B} & \textbf{30.67} & 75.73 & 83.54 &84.07 & \textbf{72.85} & 68.50 & 75.76 & 71.85 & 49.13 & 83.69  & 60.08 & 67.15 & 68.28 & 58.73\\
Swin-T & 47.7M & 64.17B & 26.96 & 77.45 & 83.67 & 83.96 & 69.49 & 69.78 & 76.54 & 73.01& 49.91 & 84.09  & 60.57 & 66.83 & 69.07 & 56.91\\
Swin-B & 88.6M & 148.3B & 9.56 & 77.92  & \textbf{85.66} &  83.31 & 70.31  & 69.14 & \textbf{77.02} & 72.94 & 52.81 & 84.75 & 61.29 &  67.85 & 69.52 & 59.02\\
ViT-S & \textbf{39.6M} & 68.53B & 35.4 & 74.65 & 81.06 &82.58 & 67.83 & 70.06 & 76.94&  71.98& 52.14 &84.16 & 60.72 & 66.87 & 69.14 & 58.89\\
DINOV2-ViT-B\dag & 105.6M & 103B &16.54 
&76.56& 83.95 & 84.95 & 71.26 & 69.58 & 75.78 & 73.46 & 51.87 & 85.96 & \textbf{61.96} & \textbf{67.98} & \textbf{70.34} &59.33\\
DINOV2-ViT-L\dag & 315.6M & 327B & 9.30 & \textbf{77.98}& 84.26 & \textbf{85.49} & 71.75 & \textbf{72.45} & 76.51 & \textbf{73.88} & \textbf{53.38} & \textbf{86.85} & 61.83 & 67.85 & 68.94 & \textbf{59.98}\\
\hline
\end{tabular}
}
\label{table:backbone_and_efficiency}
\end{table*}


\subsubsection{Hyperparameter Sensitivity Analysis}
\begin{figure}[htbp]
  \centering
  \includegraphics[width=\linewidth]{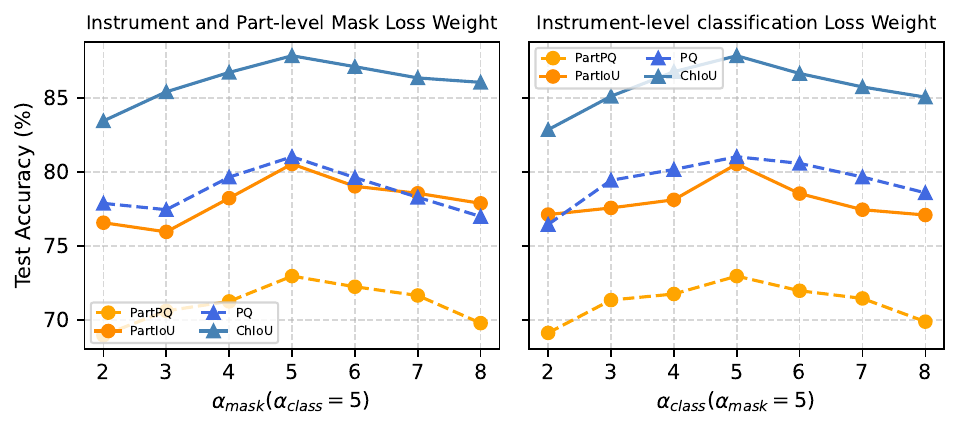}
  \caption{Sensitivity analysis for mask loss weight $\alpha_{mask}$ as discussed in \secref{sec:impdetails} with the instrument-level classification loss weight $\alpha_{class}$ fixed to 5 (left) and $\alpha_{class}$ when $\alpha_{mask}$ fixed to 5 (right), on the EndoVis2018 dataset.\label{fig:param_sensitivity_mcw}}
\end{figure}
\begin{figure}[htbp]
  \centering
  \includegraphics[width=\linewidth]{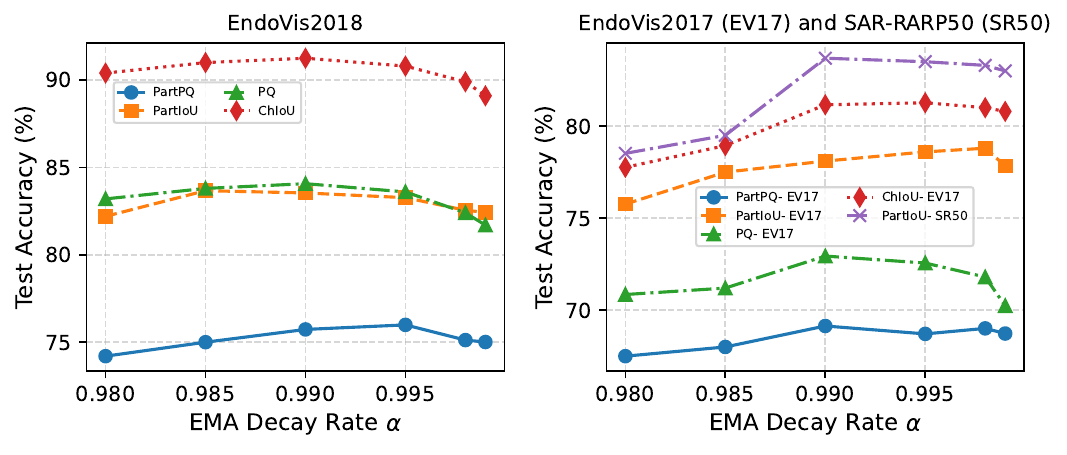}
  \caption{EMA decay rate sensitivity analysis for EndoVis2018, EndoVis2017, and SAR-RARP50 datasets.\label{fig:param_sensitivity_ema}}
\end{figure}

To select the optimal hyperparameter values to make the model achieve the highest validation accuracy, we perform a sensitivity analysis for different hyperparameter values. We perform an ablation study to assess the effect of loss weights for part-level segmentation ($\alpha_{mask}$) and instrument-level classification ($\alpha_{class}$). As shown in \figref{fig:param_sensitivity_mcw}, performance on the \gls{pis} (PartPQ), \gls{iss} (PartIoU), and \gls{iis} (PQ and Ch\_IoU) tasks displays a convex-like trend with respect to both loss weights, indicating a synergistic effect, where improvements in one task positively influence others. The results also show that setting both $\alpha_{mask}$ and $\alpha_{class}$ within the range of 4 to 6 leads to optimal model performance. Similar results are observed in \figref{fig:param_sensitivity_ema}, where an EMA decay rate value around 0.995 consistently yields optimal performance across all evaluated metrics on different tasks, regardless of the dataset.

\subsubsection{Foundation Models and Computational Efficiency vs. Task Performance}
Our approach can easily incorporate different backbones, potentially pre-trained as foundation models, to improve computational efficiency or improve task performance.
We evaluate replacing the ResNet-50 backbone with different transformer backbones in ~\tabref{table:backbone_and_efficiency} using $512 \times 512$ for the input image size.
The results highlight the trade-off between inference accuracy and computational efficiency. 
The measured inference time of our original setup with ResNet-50 achieves 30 FPS, demonstrating that our approach is already suitable for real-time applications.
Small transformers like ViT-S, when trained from scratch similarly, offer improved inference speed with moderate accuracy reduction over our ResNet-50 backbone based approach.
Larger models like Swin-B further boost accuracy but at a significant computational expense, limiting real-time usability.
To showcase the potential benefits of foundation model backbones, we incorporated ViT-B and ViT-L backbones pre-trained on LVD-142M by DINOv2 \cite{oquab2024dinov2}.
Compared to the ResNet-50 backbone, while the gain in PartPQ using DINOV2-ViT-L is +2.25 pp for EndoVis2018, it introduces significant computational overhead ($6\times$ Params), which hinders real-time applicability.

\section{Conclusion}
\label{sec:conclusion}
In this paper, we introduce SurgPIS, the first \gls{pis} model for surgical instruments. Unlike existing methods that treat instrument segmentation as separate tasks, our approach unifies instance and part segmentation using a shared-query mechanism for more structured representation learning. To overcome dataset limitations, we propose a weakly-supervised learning strategy for SurgPIS to learn from datasets that lack for \gls{pis} labels by aggregating instance and part predictions, facilitating knowledge transfer from fully labelled datasets. Extensive experiments on multiple datasets demonstrate our model’s superiority over state-of-the-art methods. Beyond \gls{pis}, our model excels in other instrument segmentation tasks, showcasing its flexibility across diverse surgical settings. Our work provides a robust foundation for structured, part-aware scene understanding in surgery, a critical step toward analyzing instrument-tissue interactions for safety monitoring and autonomous systems. In future work, we will explore the potential application for real-world robotic-assisted surgery.

\bibliographystyle{IEEEtran}
\bibliography{robo,
  ieeebib_settings
  }

\end{document}